%% file: main.tex
\documentclass[11pt,a4paper]{article}
\usepackage{acl2021}
\usepackage{times}
\usepackage{latexsym}
\usepackage{multirow}
\usepackage{multicol}
\usepackage{graphicx}
\usepackage{amsmath}
\usepackage{amssymb}
\usepackage{bbm}
\usepackage{subfigure}
\usepackage{amsfonts}
\usepackage{bm}
\usepackage{bbding}
\usepackage{booktabs}
\usepackage{color}
\usepackage{longtable}
\definecolor{sourceconcept}{RGB}{218,48,33}
\definecolor{targetconcept}{RGB}{71,159,71}
\usepackage{microtype}

\def\aclpaperid{516} 

\title{On Commonsense Cues in BERT for Solving Commonsense Tasks}
\aclfinalcopy
\author{
  Leyang Cui$^{\triangle \dag}$, Sijie Cheng$^{\ddag}$, Yu Wu$^\Diamond$, Yue Zhang$^{\dag}$\thanks{\ \ Corresponding Author} \\
  $^\triangle$ Zhejiang University \\
  $^\dag$School of Engineering, Westlake University \\
  $^\ddag$Fudan University $^\Diamond$Microsoft Research Asia \\
  cuileyang@westlake.edu.cn \ sjcheng20@fudan.edu.cn
  \\ Wu.Yu@microsoft.com \ yue.zhang@wias.org.cn
  }

\date{}

\begin{document}
\maketitle

\begin{abstract}

BERT has been used for solving commonsense tasks such as CommonsenseQA. While prior research has found that BERT does contain commonsense information to some extent, there has been work showing that pre-trained models can rely on spurious associations (e.g., data bias) rather than key cues in solving sentiment classification and other problems. We quantitatively investigate the presence of structural commonsense cues in BERT when solving commonsense tasks, and the importance of such cues for the model prediction. Using two different measures, we find that BERT does use relevant knowledge for solving the task, and the presence of commonsense knowledge is positively correlated to the model accuracy.
\end{abstract}

\section{Introduction}

Pre-trained language models \cite{elmo, gpt2, bert, roberta} give competitive results on a variety of NLP tasks \cite{bert-dependency-parsing, bert-coreference-resolution, bert-summarization, mutual}. 
It has been shown that they can effectively capture syntactic features \cite{goldberg-syntactic}, semantic information \cite{liu-probing} and factual knowledge \cite{lm-kb}, which provides support for the success in downstream tasks.

Recently, there has been some debate about whether commonsense knowledge can be learned by a language model trained on large corpora.
While \citet{commonsense-mining}, \citet{COMETCT} and \citet{cos-e} argue that pre-trained language models can directly identify commonsense facts,
\citet{kagnet} and \citet{attention-not-commonsense} believe that structured commonsense knowledge is not captured well.

Pre-trained language models have achieved empirical success when fine-tuned on specific commonsense tasks such as \textsc{Cosmos} QA \cite{cosmos}, SWAG \cite{swag}, and CommonsenseQA \cite{commonsenseqa}. One possible reason of the high performance is that there exist {\it superficial cues} or {\it spurious associations} in the dataset, which enables models to answer questions without understanding the task \cite{niven-kao-2019-probing, Yu2020ReClor, sentimentbias}. For example, a model can choose the spurious cue word ``meadow'' as a feature for positive reviews simply because ``meadow'' occurs frequently in positive documents. It remains an interesting research question whether commonsense knowledge plays a central role among statistical cues that BERT has when solving commonsense tasks. In other words, we are interested in investigating whether BERT solves commonsense tasks using commonsense knowledge.
\begin{figure}[!t]
    \centering
    \includegraphics[width=1.0\linewidth]{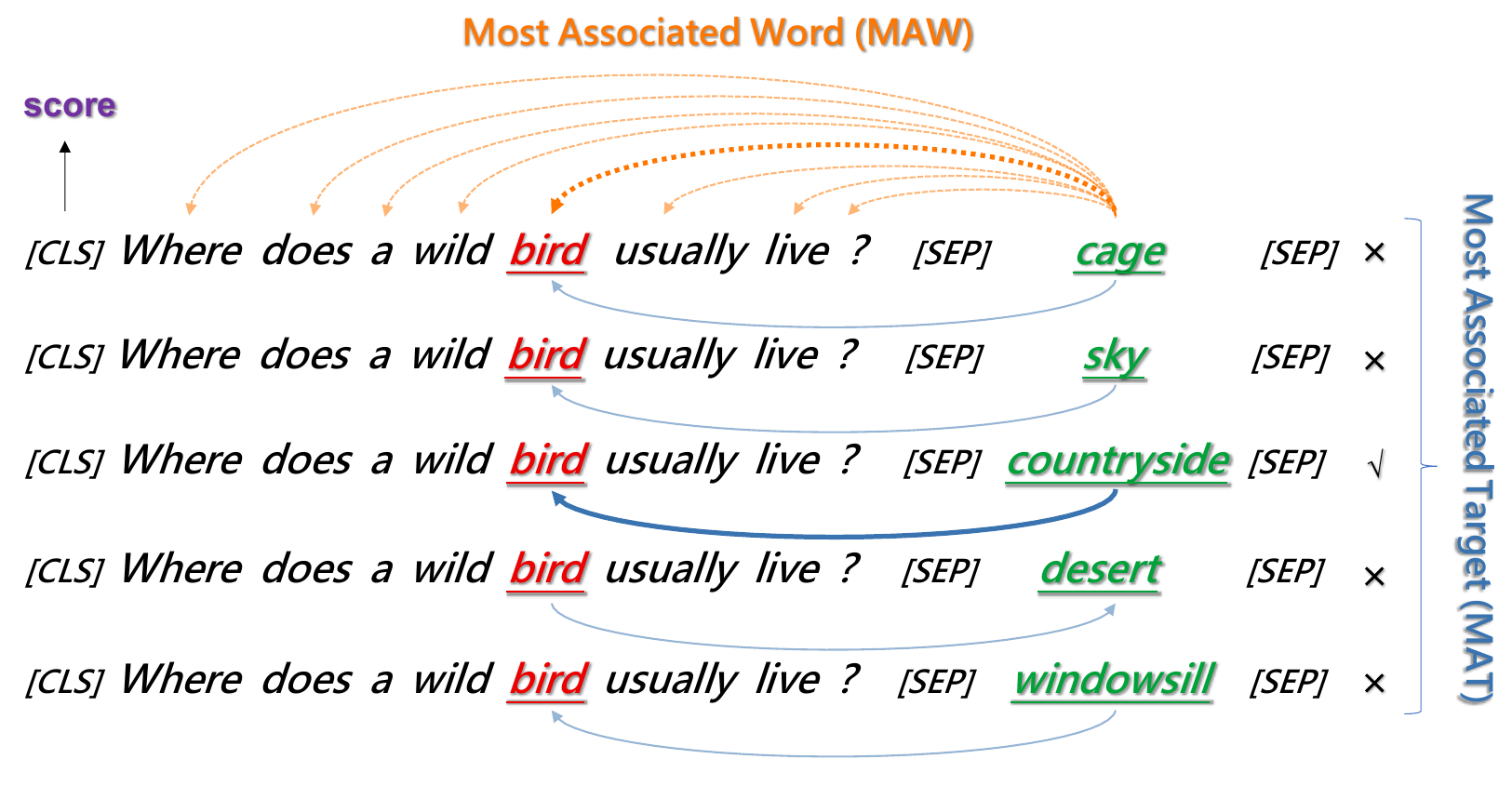}
    \caption{Two methods used to study structured commonsense knowledge in pre-trained Transformer. Commonsense link is drawn from the \textcolor{targetconcept}{\underline{Target Concept} (\underline{Answer Concept})} to the \textcolor{sourceconcept}{\underline{Source Concept} (\underline{Question Concept})}.}
    \label{fig:intro}
\end{figure}


We try to provide quantitative answers by mainly using the CommonsenseQA dataset, which asks a model to solve a multiple-choice problem. As shown in Figure~\ref{fig:intro}, given a question and five candidate answers, a model should select one candidate answer as the output. The current state-of-the-art pre-trained language models solve the problem by representing the question jointly with each candidate answer (we call such a question-answer pair a {\it sentence} thereafter), and using pre-trained language models as the main encoder. Scoring of each sentence is based on a sentence-level hidden vector, and the candidate answer that corresponds to the highest-scored sentence  is taken as the output.
 
\begin{figure}[!t]
    \centering
    \includegraphics[trim=0.1cm 5.0cm 0.1cm 4.0cm, clip=true, width=1.0\linewidth]{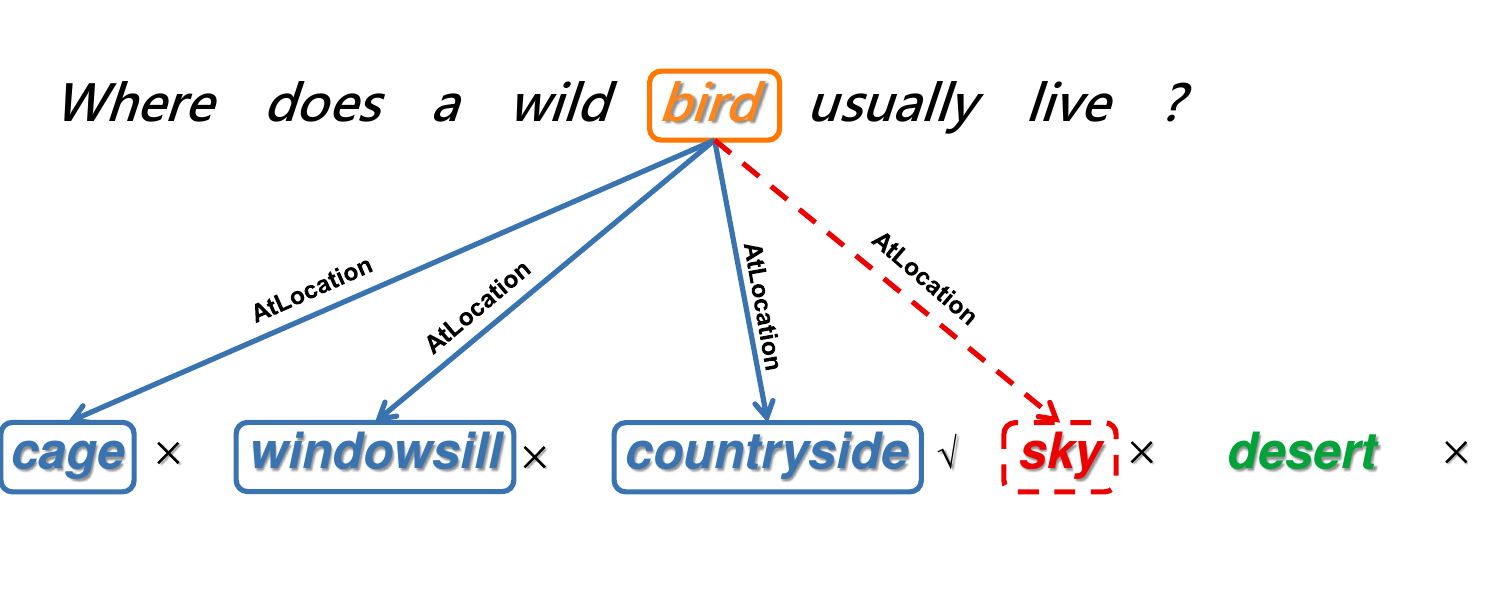}
    \caption{From \textsc{ConceptNet} to CommonsenseQA.}
    \label{fig:cqa}
\end{figure}

We investigate the presence of commonsense cues in the BERT representation of a sentence by examining {\bf commonsense links} from the {\bf answer concepts} to its related contextualized {\bf question concepts}. Figure~\ref{fig:cqa} shows one example, where the question concept is ``bird'', and the correct answer is the answer concept connected through an \textsc{AtLocation} link in the \textsc{ConceptNet} knowledge graph. Such related concepts are {\it not} explicitly used in a BERT model for making prediction, and therefore its strength in the BERT representation is not necessarily optimized in task fine-tuning. We call such cues {\it structured commonsense}, which is a source of knowledge that we can explicitly measure. We take two methods for measuring structured commonsense in BERT, including directly measuring the attention weights \cite{attention-bert} and measuring attribution scores by considering gradients \cite{attri-rc}.

We conduct two sets of experiments to quantitatively measure commonsense links in different situation. In the first set, we examine the presence of commonsense links directly in the BERT representation both before and after fine-tuning (Section~\ref{sec:method1}). In the second set of experiments, we investigate the correlation between commonsense links with model predictions (Section~\ref{sec:method2}). While the former can serve as a probing task for understanding commonsense learned by pre-training, the latter can serve as a means for understanding whether a model learns to make better use of commonsense knowledge through supervised fine-tuning.

Results suggest that BERT does have commonsense knowledge from pre-training, just as syntactic and word sense information. In addition, through fine-tuning, BERT relies more on commonsense cues in making prediction. The evidence is quantitatively demonstrated by stronger commonsense links in the representation, and a salient correlation between model predictions and commonsense link strengths, despite the fact that neither the answer concept nor the related question concept in a commonsense link is directly connected to the output layer. Interestingly, results also indicate that the stronger the structured commonsense knowledge is, the more accurate the model is. In addition to CommonsenseQA dataset, we observe similar phenomenon on Wikipedia and OMCS, demonstrating the generalization of our findings. To our knowledge, we are the first to investigate key statistical cues when BERT solves the CommonsenseQA task, providing several evidences that commonsense knowledge is indeed made use of. We release our code at \url{https://github.com/Nealcly/commonsense}.


\section{Related Work}

There has been much recent work exploiting the underlying knowledge embedded in BERT representations. \citet{elmo} find that lower layers and higher layers in ELMo contain more syntactic and semantic information, respectively.
\citet{tenney-probing}, \citet{liu-probing} and \citet{bert-structure} use probing models on hidden states to analyze linguistic information within pre-trained language models. \citet{goldberg-syntactic} assess BERT’s syntactic abilities by masking the verb, and comparing the prediction probability of the original verb with incorrect verbs. 
Our method is similar to \citet{attention-bert} and \citet{htut2019attention}, who focus on attention heads. The difference lies in that our primary goal is to investigate what information is learned and made use of when solving commonsense tasks. Therefore, our investigation is {\it task-centered}. 

There has also been work investigating data bias and spurious associations. 
\citet{nli-1} and \citet{nli-2} show that classifiers achieve accuracies around 69\% on SNLI \cite{snli} by using partial input. \citet{sentimentbias} demonstrate BERT solve sentiment analysis and NLI by heavily relying on spurious associations.
Our work is in line in investigating statistical cues. Different from the above investigations, we use probing methods to verify the presence and importance of the key feature, namely commonsense knowledge, in solving commonsense QA, rather than focusing on adversarial cases.


Commonsense reasoning is a challenging task in natural language processing. Traditional methods rely heavily on hand-crafted features \cite{wsc-feature1, wsc-feature2} and external knowledge bases \cite{wsc-knowledge}. With recent advances in deep learning, pre-trained language models have been used as a powerful method for such tasks. 
\citet{Trinh2018ASM} use a pre-trained language model to score candidate sentences for the Pronoun Disambiguation and Winograd Schema Challenge \cite{wsc-dataset}. 
\citet{contrastive} use a sentence-level loss to enhance commonsense knowledge in BERT.
\citet{commonsense-story} demonstrate that pre-trained language models fine-tuned on SWAG \cite{swag} are able to provide commonsense grounding for story generation. 
For commonsense question answering, pre-trained language models with fine-tuning give the state-of-the-art performance \cite{swag, cosmos, commonsenseqa}. Though the above work show usefulness of BERT on comonsense tasks, little work has been done investigating the mechansim for BERT solving the tasks. Our work thus complements existing research in this aspect.

There is also a line of work leveraging \textsc{ConceptNet} to enhance model's commonsense reasoning ability. \citet{kagnet} inject path information from question concepts to answer concepts to a model. \citet{commonsense-pretraining} use \textsc{ConceptNet} to construct pre-training dataset for BERT. 
\citet{lv2019graphbased} extract evidence from \textsc{ConceptNet} and Wikipedia to build a relational graph for CommonsenseQA. We use \textsc{ConceptNet} for measuring commonsense knowledge in BERT.





\section{Task and Model}
We review the main experimental dataset CommonsenseQA (Section~\ref{sec:Commonsense Linking}), before showing the structure of a state-of-the-art model (Section~\ref{sec:bert}). 

\subsection{Dataset}
\label{sec:Commonsense Linking}

CommonsenseQA \cite{commonsenseqa} is a multiple-choice question answering dataset constructed based on the knowledge graph \textsc{ConceptNet} \cite{conceptnet}, which is composed of a large set of triples taking the form $\langle$source concept, relation, target concept$\rangle$, such as $\langle$\textsc{Bird}, \textsc{AtLocation}, \textsc{Countrysides}$\rangle$.
Given a source concept \textsc{Bird} and the relation type \textsc{AtLocation}, there are three related target concepts \textsc{Cage}, \textsc{Windowsill} and \textsc{Countryside} in \textsc{ConceptNet}.

As shown in Figure~\ref{fig:cqa}, in the development of the CommonsenseQA dataset, crowd-workers are requested to generate question and candidate answers based on the {\it source concept} and three related target concepts in \textsc{ConceptNet}, respectively. Following \citet{commonsenseqa}, we call the source concept in the question as {\it question concept}, and the target concept in the answer as {\it answer concept}. Each question corresponds to only one correct {\it answer concept} among the three related \textsc{ConceptNet} target concepts. In addition, two more incorrect answer concepts are added, which do not correlate with the question concept in \textsc{ConceptNet}, resulting in 5 candidate answers for each question. We define {\it commonsene link} as the link from the answer concept to the question concept. 

The CommonsenseQA dataset is designed to avoid salient bias in surface patterns. First, the lexical overlap between the correct answer and the question is similar to that between the question and incorrect candidates. Second, commonsense links are not superficial patterns that can be learned from training data. In particular, the percentage of answer-question-concept pairs in test examples that also exist in the gold training examples is 15.78\%, which suggests that the main source of strong commonsense links, if observed, comes mainly from the pre-trained BERT model itself. 

In order to analyze implicit structured commonsense knowledge, which is based on the link from the answer concept to the question concept, 
we filter out questions which do not contain explicit mentions to the {\it question concept} in its \textsc{ConceptNet} form (e.g. paraphrase). The resulting dataset CommonsenseQA* contains 74 fewer instances.



\subsection{Model} \label{sec:bert}
We adopt the method of \citet{commonsenseqa}, using BERT \cite{bert}. 
In particular, given a question $q$ and 5 candidate answers $a_1, ..., a_5$, we concatenate the question with each answer to obtain 5 question-answer pair sequences (i.e. sentences) $s_1, \dots, s_5$, respectively. In each sentence, we use a special symbol {\tt [CLS]} in the beginning, a symbol {\tt [SEP]} between the question and the candidate answer, a symbol {\tt [SEP]} in the end.

BERT uses $L$ stacked Transformer layers \cite{transformer} to encode each sentence. The last layer hidden state of the {\tt [CLS]} token is used for linear scoring with softmax normalization. The candidate among $s_1,\dots,s_5$ with the highest score is chosen as the prediction. 
More details of our implementation are shown in Appendix~\ref{app:Implementation Details}.

\section{Analysis Methods}
As mentioned earlier, we analyze commonsense links using the attention weight \cite{attention-bert} and the corresponding attribution score \cite{ig, attri-rc}. We report results in one random execution for each experiment. We additionally tried five runs for each experiments, and found that the result variation is small (Appendix~\ref{sec:details_results}).

\subsection{Attention Weights}
\label{sec:attention}
Given a sentence, attention weights in Transformer can be viewed as the relative importance weight between each token and the other tokens when producing the next layer representation \cite{revealingBERT, interpretability-attention}.
In particular, given a sequence of input vectors $\mathbf{H} = [\mathbf{h}_1, \mathbf{h}_2,\dots, \mathbf{h}_{|H|}]$, its self-attention representation uses each vector as a query to retrieve all context vectors in $\mathbf{H}$, yielding a matrix of attention weights $\alpha \in \mathbb{R}^{|H| \times |H|}$.

The value of $\alpha$ is computed using the scaled dot-product of the query vector of representation $\mathbf{Q} = \mathbf{W}^Q \mathbf{H}$ and the key vector of representation $\mathbf{K} = \mathbf{W}^K \mathbf{H}$, followed by softmax normalization
\begin{equation}
    \alpha = softmax(\frac{\mathbf{QK}^\mathsf{T}}{\sqrt{d_k}}),
\end{equation}
where $d_k$ is the dimension size of the key vector $\mathbf{K}$. $\alpha_{i,j}$ represents the attention strength from $\mathbf{h}_i$ to $\mathbf{h}_j$.
For multi-head attention, $\mathbf{H}$ is linearly projected $T$ times to find $T$ sets of queries, keys, and values, where $T$ is the number of heads. The attention operation of each head is performed in parallel, with the results being concatenated. We use $\alpha^{m,n}$ to denote the $n$-th attention head in the $m$-th layer. The attention weights $\alpha^{m,n}$ are used as a first measure of commonsense link strengths. 

\subsection{Attribution Scores}
\label{sec:attribution}

\citet{weakAttention} point out that analyzing only attention weights can be insufficient for investigating the behavior of attention heads.
As a supplement, gradient-based feature attribution methods have been studied to interpret the contribution of each input feature to the model prediction in back-propagation \cite{attribute-classification, attri-rc, hao2020Attribution}. Analysis of both attention weights and the corresponding attribution scores allows us to more comprehensively understand commonsense links in BERT.

We employ an attribution technique called {\it Integrated Gradients} \cite{ig}. Intuitively, integrated gradients works by simulating the process of pruning the specific attention head (from the original attention weight $\alpha$ to a zero vector $\alpha'$), and calculating the integrated gradients in back-propagation. The attribution score directly reflects how much a change of attention weights affects model outputs. A higher attribution score represents more importance of the corresponding individual attention weight.
Suppose that $F(x)$ represents the BERT model output for CommonsenseQA given an input $x$. The attribution of attention head $\alpha_t t \in [1,\dots,T]$ in a Transformer layer can be computed by comparing with a set of baseline weights $\alpha'$:
\begin{equation}
Atr(\alpha^t) = (\alpha^t - \alpha'^t) \otimes \int_{x=0}^{1} \frac{\partial F(\alpha' + x (\alpha - \alpha'))}{\partial \alpha^t} dx
\end{equation}
where $\otimes$ denotes element-wise multiplication, $\alpha=[\alpha^1,\dots,\alpha^{T}]$.
Intuitively, $F(\alpha' + x(\alpha - \alpha'))$ is closer to $F(\alpha')$ when $x$ is closer to $0$, and closer to $\alpha$ when $x$ is closer to $1$. Therefore, $\int_{x=0}^{1} \frac{\partial F(\alpha' + x (\alpha - \alpha'))}{\partial \alpha^t} dx$ gives the amortized gradient with all different $x$. 
$Atr(\alpha^t) \in \mathbb{R}^{n \times n}$ denotes the attribution score which corresponding to the attention weight $\alpha^t$. 
$Atr(\alpha^t_{i,j})$ is represented for the interaction from token $\mathbf{h}_i$ to $\mathbf{h}_j$. We set the uninformative baseline $\alpha'$ as zero vector. 
Following \citet{ig}, we approximate $Atr(\alpha^t)$ via a gradient summation function,
\begin{equation}
    Atr(\alpha^t):: = (\alpha^t - \alpha'^t) \otimes  \sum_{i=1}^s \frac{\partial F(\alpha' + i/s (\alpha - \alpha'))}{\partial \alpha'^t} \times \frac{1}{s},
\end{equation}
where $s$ is the number of approximation steps for computing integrated gradients. We set $s$ to 20 based on the empirical results. 

\section{The Presence of Knowledge}
\label{sec:method1}



\begin{table*}[]
    \centering
    \small
    \begin{tabular}{c|c|c|c|c|c|c|c}
    \hline
    \multirow{2}{*}{} & \multicolumn{2}{c|}{\multirow{2}{*}{{\bf Statistics}}} & \multicolumn{4}{c|}{{\bf \textsc{Maw} Accuracy}} & \multirow{3}{*}{{\bf Random}} \\
    \cline{4-7}
     & \multicolumn{2}{c|}{} & \multicolumn{2}{c|}{{\bf Max}} & \multicolumn{2}{c|}{{\bf Avg}} &  \\
    \cline{1-7}
    Dataset & \# Instances & \# Avg Length & BERT & BERT-FT & BERT & BERT-FT & \\
    \hline
    CommonsenseQA$^*$ & 1,147 & 13.18 & 46.82 & 49.22 & 12.38 & 17.35 & 10.53 \\
    OMCS & 37,895 & 7.63 & 88.48 & 89.14 & 37.82 & 39.52 & 24.11 \\
    Wikipedia & 176,449 & 16.40 & 40.24 & 43.53 & 13.19 & 13.48 & 6.22 \\
    \hline
    \end{tabular}
    \caption{Average and maximum \textsc{Maw} accuracies across three datasets. BERT-FT model denotes the BERT model with fine-tuning on CommonsenseQA training set.}
    \label{tab:maw}
\end{table*}

\begin{table}[t!]
\small
    \centering
    \begin{tabular}{c|c|c|c|c}
        \hline
        {\bf Relation Type} & {\bf Max} & {\bf Avg} & {\bf L-H} & {\bf \# Ins}\\
        \hline
        Random & 10.53 & 10.53 & -  & -\\
        \textsc{OverAll} & 49.22 & 17.35 & 8-7 & - \\
        \hline
        \textsc{AtLocation} & 55.85 & 18.42 & 8-7 & 574 \\
        \textsc{Causes} & 55.93 & 18.91 & 8-7 & 162  \\
        \textsc{CapableOf} & 47.88 & 14.71 & 8-1 & 104 \\
        \textsc{Antonym}  & 52.53 & 10.97 & 4-3 & 83 \\
        \textsc{HasPrerequisite} & 54.15 & 18.93 & 9-8 & 41 \\
        \textsc{HasSubevent}  & 55.29 & 18.74 & 9-0 & 34 \\
        \textsc{Desires} & 40.00 & $\ \ $7.92 & 8-1 & 27 \\
        \textsc{CausesDesire} & 48.89 & 14.28 & 4-0 & 27 \\
        \textsc{PartOf} & 59.09 & 18.56 & 9-0 & 22 \\
        \textsc{HasProperty}  & 54.00 & 15.12 & 9-1 & 20 \\
        \textsc{MotivatedByGoal} & 75.56 & 24.31 & 9-7 & 18 \\
        \textsc{HasA}  & 68.89 & 22.10 & 8-1 & 9 \\
        \textsc{RelatedTo}  & 62.22 & 18.44 & 9-0 & 9 \\
        \hline
    \end{tabular}
    \caption{The average and maximum \textsc{Maw} accuracies of BERT-FT for different commonsense relations. We exclude the relation types with frequencies of occurrence less than 9. L-H represents the best performing attention head for each relation.}
    \label{tab:relation}
\end{table}

We first conduct a set of experiments to investigate commonsense link strengths in BERT representations of question-answer pairs (i.e. sentences).
Intuitively, if the link weight from the answer concept to the question concept is higher than those from the answer concept to other question words, then we have evidence of BERT using commonsense cues according to \textsc{ConceptNet}. As mentioned earlier, rather than the question concept, the representation of the {\tt [CLS]} token is directly connected to the output layer for candidate scoring. Hence there is no direct supervision signal from the output layer to the link weight during fine-tuning, and better prediction does not necessarily indicate strong commonsense links. 

\subsection{Probing Task}
Without losing generality, we call both attention weights in Section~\ref{sec:attention} and attribution weights in Section~\ref{sec:attribution} {\it link weight}.
We evaluate link weights by calculating the {\bf most associated word} (\textsc{Maw}), namely the question concept word that receives the maximum link weight from the answer concept among all question words. \textsc{Maw} is measured for each individual attention head in each layer.

Denote the hidden states of the whole question, question concept and answer concept as $[\mathbf{h}_1, \dots, \mathbf{h}_{|q|}]$,  $[\mathbf{h}_{b_s}, \dots, \mathbf{h}_{e_s}]$ and $[\mathbf{h}_{b_t}, \dots, \mathbf{h}_{e_t}]$, respectively. If the answer concept is composed of multiple tokens, we consider the link weight from the answer concept to the question token $\mathbf{h}_i$ ($i \in [1,|q|]$) as the mean of the link weights over all answer tokens $\alpha_{i} =\frac{1}{e_t - b_t} \sum_{j=b_t}^{e_t}\alpha_{j,i}$. For the $n$-th attention head in the $m$-th layer, if the question concept receives the maximum link weight from the answer concept (i.e., $\mu^{m,n}=\mathop{\arg\max}_{i} \alpha^{m,n}_{i}$, $\mu^{m,n} \in [b_s, e_s]$), we consider that this attention head gives the correct \textsc{Maw}.

We take two different measures of \textsc{Maw} accuracies, calculating the average accuracy among all attention heads, and the accuracy of the most-accurate head, respectively. Previous work probing syntactic information from attention head takes the second method \cite{attention-bert, htut2019attention}. We additionally measure the average in order to comprehensively evaluate the prevalence of commonsense cues in BERT.

The average \textsc{Maw} accuracy is measured by: 
\begin{equation*}
acc^{avg} = \frac{\sum_{m=1}^{12} \sum_{n=1}^{12} \sum_{d=1}^D \mathbbm{1}(\mu^{m,n} \in [b_s, e_s])}{12 \times 12 \times D}.
\end{equation*}

The maximum \textsc{Maw} accuracy is measured by: 
\begin{equation*}
acc^{max} = \max_{m=1}^{12} \max_{n=1}^{12}\frac{\sum_{d=1}^D \mathbbm{1}(\mu^{m,n} \in [b_s, e_s])}{D},
\end{equation*}

\noindent where $D$ represents the number of instances for evaluation.

In theory, if link weights for each attention head are randomly distributed, the average and maximum \textsc{Maw} accuracies should be both

\begin{equation*}
acc^{baseline} = \frac{\sum_{d=1}^D \frac{e_s - b_s}{|q|}}{D},
\end{equation*}

\noindent which reflects the fact that the representation does not contain explicit correlation between the answer concept and its related question concept. In contrast, \textsc{Maw} accuracies significantly better than this baseline indicates that commonsense knowledge is contained in the representation.

\subsection{Results}
\label{sec:maw_results}


The results for off-the-shelf BERT (BERT) and a BERT model fine-tuned on CommonsenseQA (BERT-FT) are shown in the first row of Table~\ref{tab:maw}. 
First, looking at the original non-fine-tuned BERT, the {\it maximum} \textsc{Maw} accuracy of each layer significantly outperforms\footnote{$p \leq 0.01$ using t-test; similar for subsequent mentions.} the random baseline.
This shows that commonsense links are a part of BERT representation of a sentence in general, just as syntactic \cite{goldberg-syntactic} and semantic \cite{liu-probing} knowledge.
Second, BERT-FT outperforms BERT in terms of both the average \textsc{Maw} accuracy and the maximum \textsc{Maw} accuracy, with a relatively large boost on the average \textsc{Maw} accuracy, which shows that structured commonsense features are enhanced by supervised training on commonsense tasks. 

We explore the best performing attentions head for each relation type in Table~\ref{tab:relation}, finding that certain attention heads capture specific commonsense relations. There is no single attention head that does well for all relation types, both with fine-tuning and without fine-tuning, which is similar to the previously finding for syntactic heads \cite{attention-mt, attention-bert}.

To further differentiate commonsense cues from superficial association, we measure the co-occurrence between each word in the question and answer concept in 1 million English Wikipedia documents. There is only 1.85\% question concept word among the highest co-occurring words of each answer concepts, which partly shows that the strong commonsense links do not heavily rely on superficial pattern.

\subsection{Additional Datasets.}
Since this set of experiments concerns the representation only, we take additional two unlabeled corpora in addition to CommonsenseQA.
In particular, we extract sentences from Open Mind Common Sense (OMCS) \footnote{Open Mind Common Sense (OMCS) corpus is the source corpus of ConceptNet.} and Wikipedia, if there existing one and only one source-target concept pair in this sentence, yielding two large-scale datasets. The detailed statistics are shown in Table~\ref{tab:maw}. The results are consistent with the CommonsenseQA dataset, which shows the generation ability of our methods. 



\begin{table}[t!]
\small
    \centering
    \setlength{\tabcolsep}{1.4mm}{
    \begin{tabular}{c|c|c|c|c|c|c}
    \hline
    & \multicolumn{4}{c|}{Attention} & \multicolumn{2}{c}{Attribution} \\
    \hline
    &
    \multicolumn{2}{c|}{\bf BERT-FT} &
    \multicolumn{2}{c|}{\bf BERT-Probing} &
    \multicolumn{2}{c}{\bf BERT-FT}
    \\
    \hline
    Head &
    \textsc{Mat} &
    \textsc{Mas} &
    \textsc{Mat} & 
    \textsc{Mas} &
    \textsc{Mat} &
    \textsc{Mas} 
    \\
    \hline
    1 &
    49.00 & 18.92 & 29.21 & \ \ 4.01 & 51.61 & 23.54
    \\
    2 &
    49.17 & 19.62 & 20.75 & 10.99 & 27.46 & 24.85
    \\
    3 &
    32.00 & 56.23 & 16.04 & 43.85 & 49.17 & 33.83
    \\
    4 &
    41.33 & 16.74 & 32.17 & \ \ 9.68 & 22.93 & 47.08
    \\
    5 & 49.96 & 24.32 & 33.91 & \ \ 6.28 & 31.04 & 44.29
    \\
    6 & 45.42 & 13.25 & 34.87 & \ \ 4.62 & 34.26 & 20.14
    \\
    7 & 48.39 & 13.33 & 25.72 & \ \ 7.41 & 33.83 & 22.67
    \\
    8 & 54.14 & 13.39 & 28.07 & \ \ 3.66 & 25.98 & 49.61
    \\
    9 &
    39.67 & 16.74 & 28.86 & \ \ 9.50 &36.97 & 22.84
    \\
    10 &
    38.71 & 13.95 & 24.50 & 18.66 & 52.14 & 21.01
    \\
    11 &
    49.17 & \ \ 8.89 & 36.88 & \ \ 7.15 & 36.79 & 21.19
    \\
    12 & 
    53.53 & 11.07 & 30.08 & \ \ 3.31 & 25.81 & 26.94
    \\
    \hline
    Avg & 
    45.87 & 18.85 & 28.42 & 10.76 & 35.67 & 29.83
    \\
    \hline
    \end{tabular}}
    \caption{$\textsc{Mat}^{overlap}$ and $\textsc{Mas}^{overlap}$ in the top layer.     \label{tab:pbt_pairs}}
\end{table}


\section{Co-relating Knowledge with Task}
\label{sec:method2}

We further conduct a set of experiments to draw the correlation between commonsense links and model prediction. The goal is to investigate 
how BERT makes use of commonsense knowledge for making a prediction in the CommonsenseQA task.
In particular, we compare the link weights across the five answer candidates for the same question, and find out the candidate that is the most associated with the relevant question concept. This candidate is called the {\bf most associated target} (\textsc{Mat}). Correlations are drawn between \textsc{Mat}s and the model prediction for each question. Intuitively, the more the \textsc{Mat}s are correlated with the model predictions, the more evidence we have that the model makes use of commonsense cues in making prediction. 


Both attention weights and the corresponding attribution scores are used, because now we are considering model prediction, for which gradients play a role and can be measured. For all experiments, the trend of attribute scores is consistent with that measured using attention weights. 

\subsection{Probing Tasks}

Formally, given a question $q$ and 5 candidate answers $a_1,\dots,a_5$, we make comparisons across five candidate sentences $s_1,\dots,s_5$. In each candidate sentence, we calculate the link weight from the answer concept to the question concept according to \textsc{ConceptNet}. Denote the hidden states of the question concept and the answer concept as $[\mathbf{h}_{b_s},\dots,\mathbf{h}_{e_s}]$ and $[\mathbf{h}_{b_t},\dots,\mathbf{h}_{e_t}]$, respectively.
The link weight of the answer-question-concept pair ($\alpha_{a2q}$) is the average between each answer concept token and each question concept token
\begin{equation*}
    \alpha_{a2q} = \frac{\sum_{i=b_s}^{e_s} \sum_{j=b_t}^{e_t} \alpha_{j,i}}{(e_s-b_s)(e_t-b_t)}
\end{equation*}

% ...
Among the five candidates in each instance, we take the one with the highest $\alpha_{a2q}$ as the most associated target \textsc{Mat}, denoted as $s^{\textsc{Mat}} \in [1,5]$.

As a baseline for \textsc{Mat}, we further define {\bf most associated sentence} (\textsc{Mas}) as the candidate answer that has the maximum link weight from the answer concept to the {\tt [CLS]} token among the five candidates. The reason is that gradients are back-propagated from the {\tt [CLS]} token rather than the question concept or the answer concept. By comparing \textsc{Mat} and \textsc{Mas}, we can have useful information on whether \textsc{Mat} is an influencing factor for the model decision. 
% ...

We measure the correlation between \textsc{Mat} ($s^{\textsc{Mat}} \in [1,5]$), the model prediction ($s^{model} \in [1,5]$) and the gold-standard answer ($s^{golden} \in [1,5]$) by using two metrics, including the overlapping rate between \textsc{Mat}s and model predictions, and the accuracy of \textsc{Mat}s. 

The {\bf overlapping rate of \textsc{Mat}s} is defined as: 
\begin{equation*}
\textsc{Mat}^{overlap} = \frac{\sum_{d=1}^D \mathbbm{1}(s^{\textsc{Mat}}_d = s^{model}_d)}{D}
\end{equation*}

The {\bf accuracy of \textsc{Mat}s} is defined as the percentage of Mats that equals the gold answer: 
\begin{equation*}
\textsc{Mat}^{acc} = \frac{\sum_{d=1}^D \mathbbm{1}(s^{\textsc{Mat}}_d = s^{golden}_d)}{D}
\end{equation*}
Similar to \textsc{Maw}, \textsc{Mat} and \textsc{Mas} can be measured for each attention head, and we calculate the average and maximum values across different heads.

\subsection{Commonsense Link and Model Output}

\begin{figure*}[t!]
    \centering
    \includegraphics[width=\textwidth]{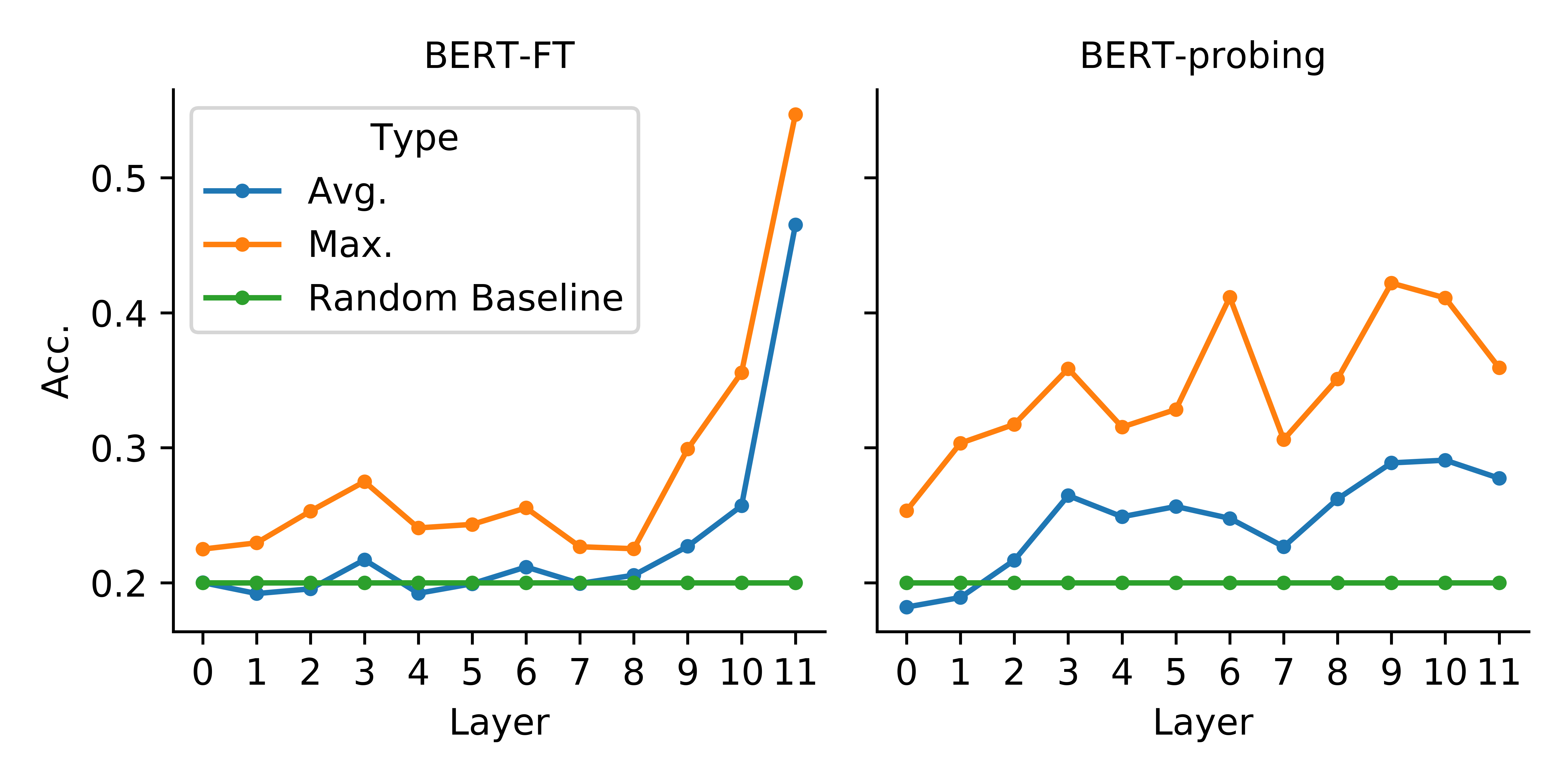}
    \caption{$\textsc{Mat}^{overlap}$ across different layers.}
    \label{fig:pat_across_layer}
\end{figure*}


\begin{table}[t!]
\small
    \centering
    \subfigure[Measured by attention weights.]{
        \begin{tabular}{c|c|c|c|c|c}
         \toprule
         {\bf \#H} & {\bf \#Ins} & {\bf Model Acc.} & {\bf \#H} & {\bf \#Ins} & {\bf Model Acc.}\\
         \midrule
         0 & 158 & 20.89 & 7 & 69 & 78.26 \\
         1 & 135 & 28.15 & 8 & 63 & 82.54 \\
         2 & 119 & 52.10 & 9 & 57 & 92.98 \\
         3 & 132 & 53.79 & 10 & 47 & 89.36 \\
         4 & 93 & 62.37 & 11 & 44 & 97.73 \\
         5 & 106 & 66.04 & 12 & 36 & 100.00\\
         6 & 88 & 68.18 & - & - & - \\
         \bottomrule
    \end{tabular}
    }
    \subfigure[Measured by attribution.]{
    \begin{tabular}{c|c|c|c|c|c}
         \toprule
         {\bf \#H} & {\bf \#Ins} & {\bf Model Acc.} & {\bf \#H} & {\bf \#Ins} & {\bf Model Acc.}\\
         \midrule
         0 & 89 & 10.11 & 5 & 171 & 72.51 \\
         1 & 114 & 22.81 & 6 & 119 & 81.51 \\
         2 & 148 & 51.35 & 7 & 85 & 82.35 \\
         3 & 156 & 56.41 & 8 & 43 & 74.72 \\
         4 & 207 & 66.67 & 9 & 13 & 84.62 \\
         \bottomrule
    \end{tabular}
    }
        \caption{The relationship between the \textsc{Mat} head count and the model prediction accuracy. \#H denotes how many heads yield the correct \textsc{Mat} prediction. \#Ins denotes the instance number. \label{tab:Mat_vs_acc}}
\end{table}


We measure the \textsc{Mat} performance of BERT-FT, and a BERT model that is fine-tuned for the output layer only (BERT-probing). The latter is a linear probing model \cite{liu-probing}. Intuitively, if the probing model can solve the commonsense task accurately, then the original non-fine-tuned BERT likely encodes the rich commonsense knowledge. 

Table~\ref{tab:pbt_pairs} shows the relative strengths of \textsc{Mat}s and \textsc{Mas}s according to the 12 attention heads in the top Transformer layer. First, for both models, the overlapping rates of \textsc{Mat}s are significantly ($p \leq 0.01$) larger than that with \textsc{Mas}s. This suggests that the link weight from the answer concept to the question concept is more closely-related to the model prediction as compared to the link weight from the answer concept to the {\tt [CLS]} token, despite that model output scores are calculated on the {\tt [CLS]} token. 
The results give strong evidence that commonsense cues from BERT are relied on for model decision.
Second, when fine-tuned with training data, the model gives an even stronger correlation between \textsc{Mat} and the model prediction.
This suggests that the model can {\it learn} to make use of commonsense cues for making prediction, which partly shows how a BERT model solves CommonsenseQA.

Figure~\ref{fig:pat_across_layer} shows the overlapping rate between \textsc{Mat} and model prediction at each Transformer layer. Both the maximum and the average overlapping rates across the 12 layers are shown. The random overlapping rate of 20\% is drawn as a reference. It can be seen from the figure that the maximum overlapping rate of BERT-probing is significantly larger than the random baseline, which shows that the model prediction is associated with the relevant structured commonsense cues. In addition, after fine-tuning, the BERT-FT model shows a tendency of weakened maximum \textsc{Mat} overlapping rate on lower Transformer layers and much strengthened \textsc{Mat} overlapping rate on higher layers, and in particular the top layer. The trend of \textsc{Mat} measured by attribution score is consistent with attention weights. This suggests that fine-tuned model relies more on the commonsense structure in the top layer for making prediction. 

We compare the co-occurrence between question concepts and candidate answer concepts in 1 million English Wikipedia documents, and find that only 18.2\% gold answers has the most co-occurrence with the question concept among 5 answer candidates, which is even lower than the random baseline (20\%), showing that CommonsenseQA cannot be solved by solely relying on superficial patterns.

\subsection{Commonsense Link and Model Accuracy}

\begin{figure}
    \centering
    \includegraphics[width=0.5\textwidth]{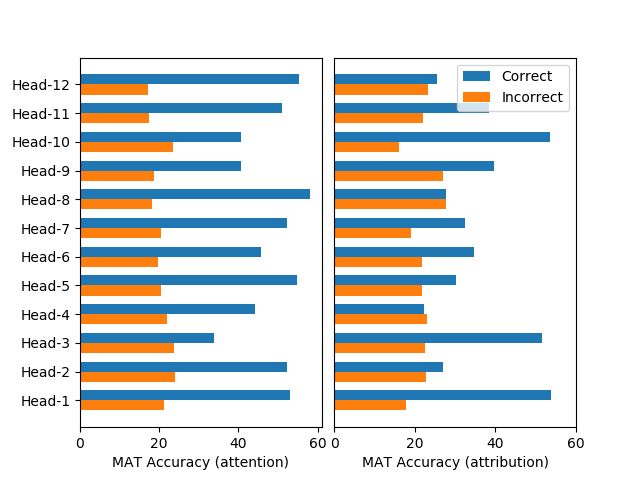}
    \caption{$\textsc{Mat}^{acc}$ of each attention head in the top layer with correct and incorrect model predictions. ``Red'' and ``Blue'' indicate the model performance if attention head-$n$ gives correct and incorrect prediction, respectively.}
    \label{fig:golden_label}
\end{figure}

Table~\ref{tab:Mat_vs_acc} shows the correlation between \textsc{Mat} accuracies and model prediction accuracies. Each row shows a different number of heads in the top layer for which the \textsc{Mat} corresponds to the correct answer candidate, together with the number of test instances for such cases, and the model prediction accuracy on the instances. There is an obvious trend where increased \textsc{Mat} accuracies correspond to increased model prediction accuracies, which shows that making use of structured commonsense cues leads to better model prediction.

Figure~\ref{fig:golden_label} shows the \textsc{Mat} accuracies of each attention head in the top layer for the test instances with correct and incorrect model predictions, respectively. The \textsc{Mat} accuracies of correctly predicted instances are larger than those of incorrectly predicted instances by a large margin. The finding is consistent with Table~\ref{tab:Mat_vs_acc}, which shows that structured commonsense cues are a key factor in BERT making the correct decision.

\begin{figure}[t!]
    \centering
    \includegraphics[width=0.5\textwidth]{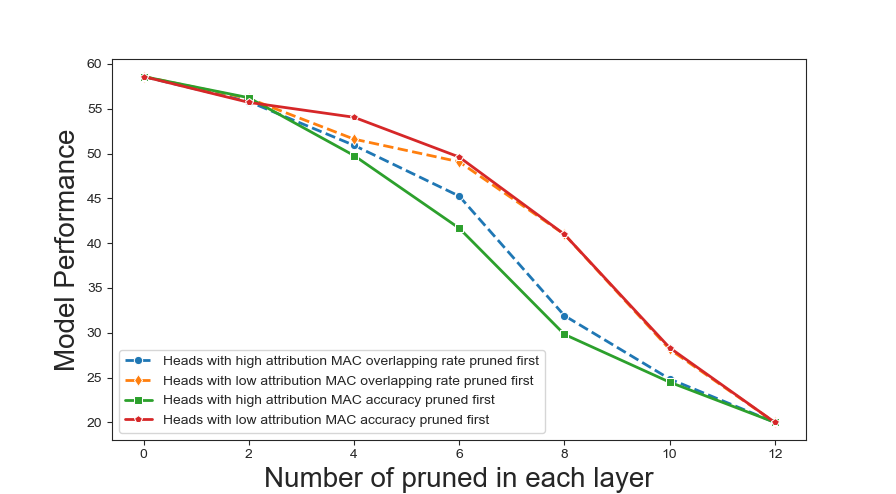}
    \caption{Model performance on the CommonsenseQA development set when different heads are pruned. 
    }
    \label{fig:pruning}
\end{figure}

\begin{table}[t!]
\small
    \centering
    \begin{tabular}{c|c|c|c|c|c|c}
    \hline
     & 
     \multicolumn{3}{c|}{\bf BERT-FT} &
     \multicolumn{3}{c}{\bf BERT-probing} \\
    \hline
     & 
    \multicolumn{2}{c|}{$\textsc{Mat}^{overlap}$} & Model &
    \multicolumn{2}{c|}{$\textsc{Mat}^{overlap}$} & Model \\
    \hline
    L & Max & Avg & Acc & Max & Avg & Acc \\ 
    \hline
    12 & 
    54.14 & 45.87 & 58.59 & 36.88 & 28.42 & 39.23 
    \\
    11 & 
    46.56 & 26.65 & 56.50 & 37.66 & 27.11 & 35.48
    \\
    10 & 
    37.40 & 27.86 & 53.36 & 39.84 & 28.50 & 33.74
    \\
    9 & 
    34.61 & 24.01 & 51.53 & 30.08 & 24.76 & 32.52 
    \\
    8 &
    31.82 & 21.39 & 49.35 & 25.81 & 21.53 & 33.57
    \\
    7 &
    31.73 & 24.40 & 48.74 & 37.05 & 24.04 & 32.96
    \\
    6 &
    31.56 & 23.64 & 45.95 & 31.21 & 24.02 & 32.00
    \\
    5 &
    34.44 & 25.01 & 44.99 & 33.39 & 24.03 & 32.43
    \\
    4 &
    44.73 & 34.13 & 40.28 & 41.06 & 27.67 & 33.83
    \\
    3 &
    44.20 & 32.48 & 37.58 & 25.81 & 21.02 & 21.88
    \\
    2 &
    23.71 & 19.47 & 26.68 & 23.63 & 20.74 & 20.40
    \\
    1 &
    23.45 & 19.50 & 23.02 & 20.58 & 18.81 & 19.27
    \\
    \hline
    \end{tabular}
    \caption{Performance of $\textsc{Mat}^{overlap}$ across different layers. L-$n$ represents adding the output classifier on the hidden state of layer-$n$. Our BERT-FT model (layer-11) gives 58.15\% accuraies, which is slightly higher than the reported results of 55.57\% on \citet{kagnet}. It achieves 58.59\% on our dataset CommonsenseQA*.}
    \label{tab:ft_on_diff_layer}
\end{table}

We further evaluate the model performance after pruning specific heads. We sort all the attention heads in each layer according to their \textsc{Mat} performance by attribution scores, and then prune these heads in order. Following \citet{pruning}, we replace the pruned head with zero vectors. Figure~\ref{fig:pruning} shows the model performance on the development set. As the number of pruned heads increases, the model performance decreases, which conforms to intuition. In addition, the model performance drops much more rapidly when the attention heads with higher \textsc{Mat} performances are pruned first, which demonstrates that capturing commonsense features is crucial to strong model prediction.

\subsection{Commonsense Link and BERT Layer}
We further investigate two specific questions on the commonsense knowledge usage. First, which layer does BERT rely on the most for making its decision. Second, does the commonsense knowledge that BERT uses come more from pre-training or fine-tuning.
We compare 12 model variations by connecting the output layer on each of the Transformer layer, respectively. Table~\ref{tab:ft_on_diff_layer} shows the model accuracies and the \textsc{Mat} overlapping rates. First, BERT-probing gives the best performance when prediction is made on the top layer, and the accuracy generally decreases as the layer moves to the bottom. This indicates that relevant commonsense knowledge is more heavily distributed towards higher layers during pre-training. Our experimental settings here are the same as the probing task for syntactic information by \citet{liu-probing}, who find that syntactic information is distributed more heavily towards lower BERT layers. 

With fine-tuning, we observe stronger improvements of both model accuracies and \textsc{Mat} overlaps on higher layers when comparing BERT-FT and BERT-probing. This demonstrates that commonsense knowledge on higher layers is more useful to the CommonsenseQA task. Interestingly, comparing layer 11 and layer 10, the model accuracy after fine-tuning is similar, but the \textsc{Mat} overlap of layer 11 is significantly larger. This can suggest that the structured commonsense knowledge that we probe attributes only partly to the overall useful knowledge for CommonsenseQA.



\section{Conclusion}
We conducted quantitative analysis to investigate how BERT solves the CommonsenseQA task, aiming to gain evidence on the key source of information involved in the disambiguation process.
Empirical results demonstrated that BERT encodes structured commonsense knowledge, and is able to leverage such cues on the downstream CommonsenseQA task. Our analysis has further revealed that with fine-tuning, BERT learns to make better use of commonsense features on higher layers.
These suggest that BERT can learn to make use of truly relevant commonsense cues rather than superficial patterns for CommonsenseQA.

\bibliographystyle{acl_natbib}
\bibliography{acl2021}

\input{appendix}

\end{document}

%% file: appendix.tex
\clearpage

\begin{appendix}

\section{Detailed performance of \textsc{Maw}}
\label{appendix:Mat_attri}
We report the average and maximum \textsc{Maw} accuracy across different layers in Table~\ref{fig:maw_across_layer}.
The average \textsc{Maw} of 6 layers significantly outperforms the random baseline, which indicates that the relevant question concept plays a highly important role in BERT encoding without fine-tuning. BERT-FT outperforms BERT in terms of both average \textsc{Maw} accuracy and maximum \textsc{Maw} accuracy, which shows that structured commonsense knowledge is enhanced by supervised training on commonsense tasks. 
\begin{table}[h!]
\small
    \centering
    \begin{tabular}{c|c|c|c|c|c|c|c}
    \hline
        & \multicolumn{3}{c|}{\bf BERT-FT} & \multicolumn{3}{c|}{\bf BERT} & \\
        \hline
        L & Max & Avg & t & Max & Avg & t & Rand \\
         \hline
        11 & 34.11 & 19.78 & \checkmark & 32.44 & 14.47 & \checkmark & 10.53  \\
        10 & 39.09 & 26.10 & \checkmark & 40.84 & 22.22 & \checkmark & 10.53 \\
        9  & 46.31 & 25.59 & \checkmark & {\bf 46.82} & 18.68 & \checkmark & 10.53 \\
        8  & {\bf 49.22} & 13.71 & \checkmark & 44.48 & 10.15 & - & 10.53 \\
        7  & 32.76 & 8.88 & - & 28.00 & 5.61 & - & 10.53 \\
        6  & 40.68 & 12.16 & \checkmark & 41.99 & 9.01 & -  & 10.53 \\
        5  & 33.30 & 14.41 & \checkmark & 13.22 & 4.34 & - & 10.53 \\
        4  & 38.89 & 19.09 & \checkmark & 24.10 & 10.46 & - & 10.53 \\
        3  & 37.30 & 14.59 & \checkmark & 24.74 & 7.43 & - &10.53 \\
        2  & 35.08 & 17.71 & \checkmark & 31.96 & 12.14 & \checkmark & 10.53\\
        1  & 29.01 & 15.08 & \checkmark & 27.64 & 11.09 & \checkmark & 10.53 \\
        0  & 45.55 & 23.05 & \checkmark & 46.16 & 22.95 & \checkmark & 10.53\\
        \hline
        All&49.22&17.35&\checkmark&46.82&12.38&\checkmark&10.53\\
        \hline
    \end{tabular}
    \caption{The average and maximum \textsc{Maw} accuracy across different layers. \checkmark indicates $p$-value \textless 0.01.}
    \label{fig:maw_across_layer}
\end{table}

\section{Performance of \textsc{Mat}}
\label{sec:details_results}

\begin{table*}[t!]
    \centering
    \small
    \begin{tabular}{c|c|c|c|c|c|c|c|c|c|c|c|c}
    \hline
        &  \multicolumn{6}{c|}{\textsc{Mat}} & \multicolumn{6}{c}{\textsc{Mas}} \\
        \hline
        Layer & M1 & M2 & M3 & M4 & M5 & mean$\pm$std & M1 & M2 & M3 & M4 & M5 & mean$\pm$std\\
        \hline
        0 & 20.10 & 20.26 & 20.32 & 20.35 & 20.58 & 20.32$\pm$0.17 & 18.86 & 18.38 & 18.40 & 18.07 & 18.19 & 18.38$\pm$0.30\\
        1 & 19.35 & 19.36 & 19.11 & 19.39 & 19.51 & 19.34$\pm$0.15 & 21.00 & 20.09 & 20.37 & 20.16 & 20.23 & 20.37$\pm$0.37\\
        2 & 20.25 & 19.57 & 19.97 & 20.13 & 19.85 & 19.95$\pm$0.26 & 19.20 & 19.61 & 20.19 & 19.03 & 19.01 & 19.41$\pm$0.50\\
        3 & 21.98 & 21.15 & 21.80 & 22.05 & 23.16 & 22.03$\pm$0.73 & 18.88 & 19.46 & 18.39 & 18.12 & 19.01 & 18.77$\pm$0.53\\
        4 & 19.78 & 19.19 & 19.91 & 20.15 & 19.27 & 19.66$\pm$0.42 & 19.99 & 19.50 & 20.05 & 22.28 & 20.76 & 20.52$\pm$1.08\\
        5 & 19.90 & 19.74 & 19.89 & 20.33 & 20.55 & 20.08$\pm$0.34 & 16.98 & 16.90 & 17.90 & 17.56 & 17.15 & 17.30$\pm$0.42\\
        6 & 19.29 & 19.31 & 19.02 & 19.25 & 18.51 & 19.08$\pm$0.34 & 15.77 & 16.25 & 15.72 & 18.27 & 17.15 & 16.63$\pm$1.08\\
        7 & 20.71 & 20.47 & 19.92 & 20.14 & 19.30 & 20.11$\pm$0.54 & 18.40 & 19.70 & 16.57 & 22.09 & 20.66 & 19.48$\pm$2.11\\
        8 & 19.48 & 21.20 & 21.00 & 20.65 & 19.13 & 20.29$\pm$0.93 & 18.16 & 19.70 & 17.28 & 21.89 & 19.12 & 19.23$\pm$1.75\\
        9 & 21.95 & 21.85 & 22.87 & 21.80 & 22.88 & 22.27$\pm$0.55 & 12.54 & 14.94 & 13.07 & 14.85 & 13.75 & 13.83$\pm$1.06\\
        10 & 25.59 & 24.95 & 25.61 & 24.96 & 25.07 & 25.24$\pm$0.34 & 16.60 & 17.80 & 15.28 & 18.12 & 16.68 & 16.90$\pm$1.12\\
        11 & 45.91 & 45.87 & 44.61 & 42.28 & 41.88 & 44.11$\pm$1.93 & 10.32 & 18.21 & $ \ \ $6.04 & 22.73 & 22.80 & 16.02$\pm$7.55\\
        \hline
    \end{tabular}
    \caption{\textsc{Mat} and \textsc{Mas} overlapping rate for each attention head across five models, as well as their average value with a standard deviation. M - Model. \label{table:appendex1}}
\end{table*}

Table~\ref{table:appendex1} shows the \textsc{Mat} and \textsc{Mas} performance for each attention head across five turns. Noted that the standard derivations are only 1.17\% and 1.76\% for \textsc{Mat} and \textsc{Mas}, respectively, which demonstrates the robustness of our methods.

\section{Implementation Details}
\label{app:Implementation Details}
We adopt the huggingface BERT-base implementation for multiple-choice on CommonsenseQA. We conduct fine-tuning experiments using GeForce GTX 2080Ti. For BERT-FT and BERT-probing, we optimize the parameters with grid search: training epochs 3, learning rate $\{5e-4, 3e-5, 5e-5, 5e-6\}$, training batch size $\{8,16,32\}$, gradient accumulation steps $\{2,4,8\}$. To demonstrate the robustness of our analysis method, we repeat the experiment 5 times with the same hyperparameter, and report the experiment results based on one random model .

We calculate the attribution score to interpret BERT using captum, which is an extensible library for model interpret ability built on Pytorch.

\end{appendix}